\documentclass[journal,twoside,web]{ieeecolor}
\usepackage{jsen}
\usepackage{cite}
\usepackage{amsmath,amssymb,amsfonts}
\usepackage{algorithmic}
\usepackage{graphicx}
\usepackage{textcomp}
\usepackage{wrapfig}
\usepackage{url}

\usepackage{multirow}
\usepackage{makecell}
\usepackage{booktabs}   
\usepackage{graphicx}

\def\BibTeX{{\rm B\kern-.05em{\sc i\kern-.025em b}\kern-.08em
    T\kern-.1667em\lower.7ex\hbox{E}\kern-.125emX}}
\definecolor{abstractbg}{rgb}{0.89804,0.94510,0.83137}
\begin{document}
\title{DV-VLN: Dual Verification for Reliable LLM-Based Vision-and-Language Navigation}
\author{Zijun Li, Shijie Li, Zhenxi Zhang, Bin Li$^*$, and Shoujun Zhou$^*$
\thanks{
This work was supported by the Shenzhen Medical Research Fund (No. D2404001), and in part by the Key Research and Development Program of Guangdong Province (No. 2025B1111020001), the Shenzhen Municipal STIB Key programs (No. CJGJZD20230724093303007, and KJZD20240903101259001), National Key Laboratory of the CAS on Medical Imaging Science and Technology System, the Xisike Clinical Oncology Research Foundation (Y-2024AZ(NSCLC)MS-0156), and SIAT-WUXI Joint Innov-Group for AGI-MET.}
\thanks{$^*$Corresponding authors: Bin Li (e-mail: b.li2@siat.ac.cn) and Shoujun Zhou (e-mail: sj.zhou@siat.ac.cn).
}
\thanks{Zijun Li is with the Robotics Engineering Program, College of Engineering, Zhejiang Normal University, Jinhua, China (e-mail: 19548990199@163.com).}
\thanks{Shijie Li is with the Shenzhen Institutes of Advanced Technology (SIAT), Chinese Academy of Sciences, Shenzhen, China (e-mail: sj.li5@siat.ac.cn).}
\thanks{Zhenxi Zhang is with the Department of Health Technology and Informatics, The Hong Kong Polytechnic University, Hong Kong (e-mail: 25112966r@connect.polyu.hk).}
\thanks{Bin Li is with the Shenzhen Institutes of Advanced Technology (SIAT), Chinese Academy of Sciences, Shenzhen, China (e-mail: b.li2@siat.ac.cn).}
\thanks{Shoujun Zhou is with the Shenzhen Institutes of Advanced Technology (SIAT), Chinese Academy of Sciences, Shenzhen, China (e-mail: sj.zhou@siat.ac.cn).}
}

\IEEEtitleabstractindextext{%
\fcolorbox{abstractbg}{abstractbg}{%
\begin{minipage}{\textwidth}%
\begin{wrapfigure}[12]{r}{3in}%
\hspace*{-10pt}
\includegraphics[width=3in]{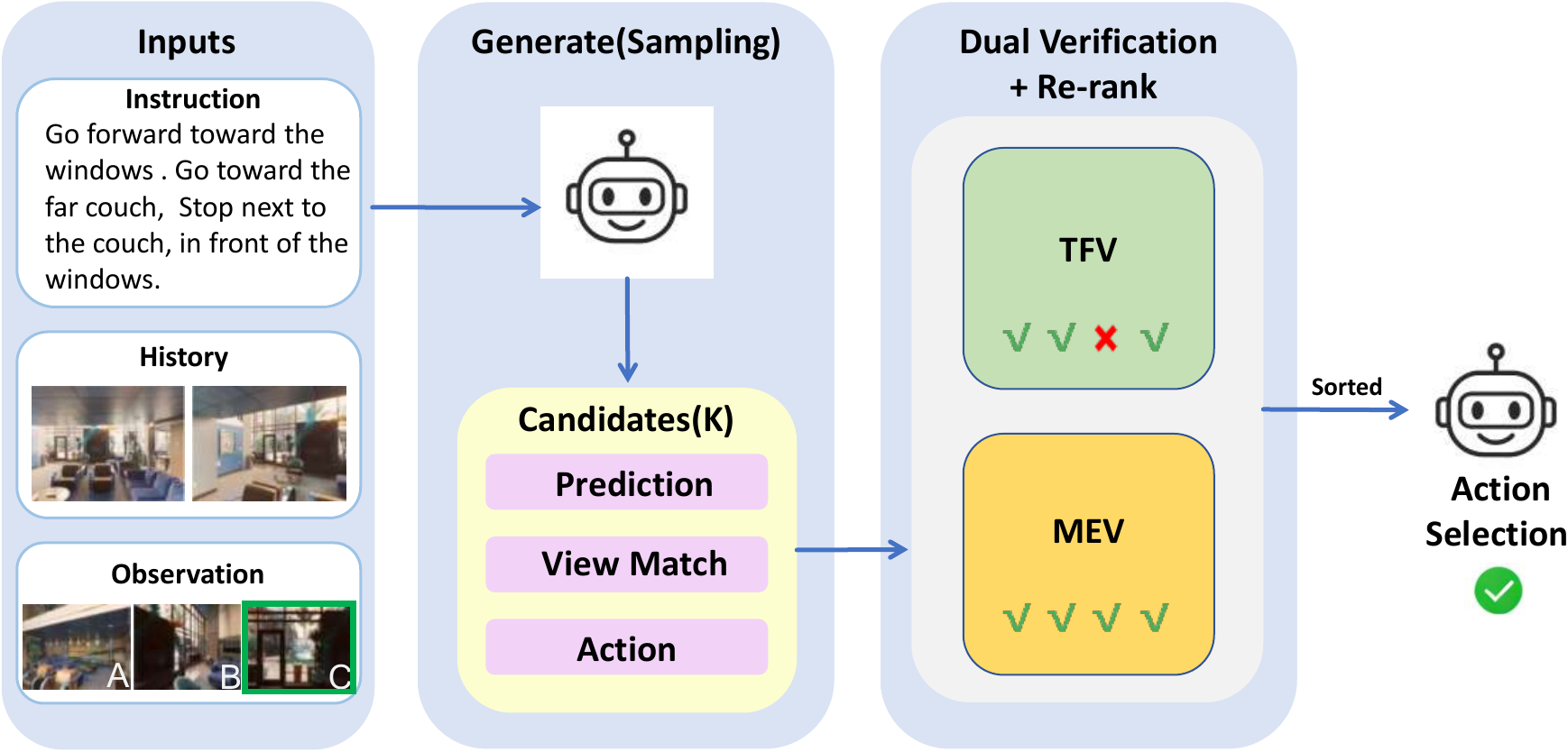}%
\end{wrapfigure}%
\begin{abstract}
Vision-and-Language Navigation (VLN) requires an embodied agent to navigate in a complex 3D environment according to natural language instructions.  The latest progress of large language models (LLMs) makes language-driven navigation more interpretable. However, most LLM-based agents still rely on a single-action decision-making, where the model must choose an option from noisy, textualized multi-perspective observations. Due to local mismatches or imperfect intermediate reasoning, these decisions can easily deviate from the correct path, resulting in the accumulation of errors and reduced reliability in unseen environments. In this paper, we propose DV-VLN, a verification-guided VLN framework that follows a generate–then–verify paradigm. DV-VLN first performs parameter-efficient in-domain adaptation of an open-source LLaMA-2 backbone to produce a structured navigational chain-of-thought in a unified format—Prediction, View Match, and Action—and then samples multiple candidate actions via sampling decoding. In order to improve reliability in reasoning, we introduce dual verification: True–False Verification (TFV) checks whether a candidate conforms to the given instructions, history and observations, while Masked-Entity Verification (MEV) masks key instruction entities and tests whether they can be recovered assuming the candidate action is taken. DV-VLN selects actions by summing verification successes across samples, yielding interpretable scores for re-ranking. Experiments on R2R, RxR (English subset), and REVERIE show that DV-VLN significantly improves over direct prediction and sampling-only baselines, delivering competitive performance among language-only VLN agents under standard training protocols and promising results compared with several cross-modal systems. Code is available at \url{https://github.com/PlumJun/DV-VLN}.
\end{abstract}

\begin{IEEEkeywords}
Vision-and-language navigation, large language models, dual verification
\end{IEEEkeywords}
\end{minipage}}}

\maketitle

\section{Introduction}
\label{sec:introduction}
\IEEEPARstart{V}{ision-and-Language} (VLN)~\cite{anderson2018vision,qi2020reverie,Chen2019TOUCHDOWNNL,jain2019stay,ku2020room} requires an embodied agent to follow the natural language instructions to reach the target position in a 3D environment. As a representative task of Embodied AI, VLN is attractive in the fields of service robots and indoor assistance, but it is also extremely challenging: the agent must combine long and complex instructions in visual observations, maintain a consistent belief over navigation history, and make continuous decisions under partial observability. Small errors in intermediate steps are easy to accumulate, causing the agent to deviate from the expected route.

The recent development of large language models (LLMs)~\cite{brown2020language,touvron2023llama,touvron2023llama2} has opened up a new path for VLN. Trained on massive corpora, LLMs exhibit strong capabilities in planning, reasoning and reflection, and have been applied to a variety of embodied tasks~\cite{Ahn2022DoAI,Huang2022InnerME,driess2023palme}. By combining the off-the-shelf vision-to-text systems~\cite{li2023blip2,li2024towards,li2022blip} with LLMs, several works convert panoramic observations into text descriptions and allow an LLM to choose actions directly from language inputs~\cite{zhou2023navgpt,long2023discuss}. Although these LLM-based agents reveal the potential of leveraging world knowledge and generating human-readable rationales, they also expose several limitations: (i) many methods rely on large closed-source models such as GPT-4~\cite{OpenAI_2023}, which are costly and difficult to deploy; (ii) there is often a substantial domain gap between internet text and VLN trajectories; and, crucially, (iii) action decisions are usually produced in a single forward pass, without any explicit mechanism to check or correct potentially wrong predictions.

\begin{figure*}[t]
\begin{centering}
\includegraphics[width=0.95\linewidth]{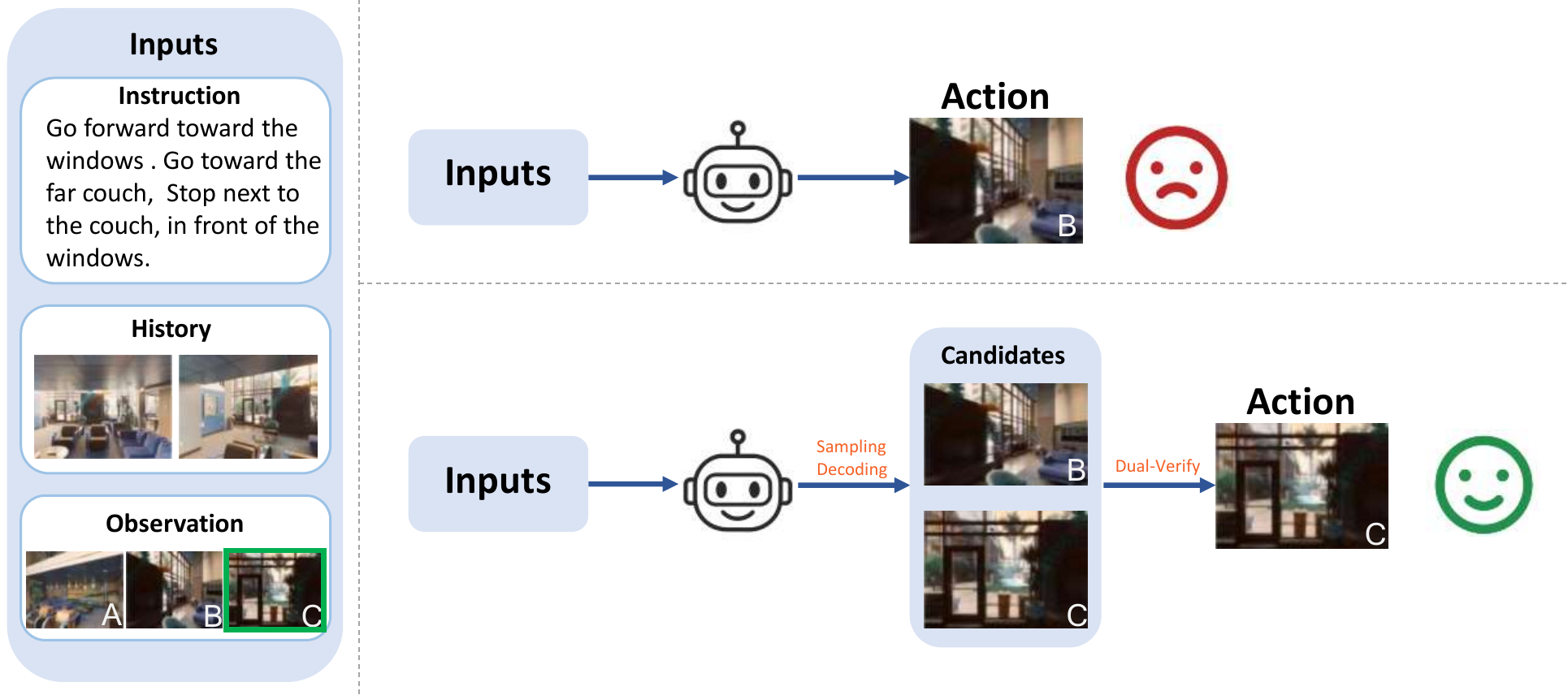}
\par\end{centering}

\caption{Comparison between direct action decision and our DV-VLN. Given the same instruction, history, and current observations, the direct LLM agent takes a single-step decision and wrongly selects view B. In contrast, DV-VLN first samples multiple candidate actions (B and C) and then applies dual verification, finally choosing candidate C that best matches the instruction and leads to the correct action.
}
\vspace{-0.6cm}
\label{fig:motivation}
\end{figure*}

As shown in Fig.~\ref{fig:motivation}, this “single-shot” decision pattern is easy to fail when observations are cluttered or there are multiple possible perspectives. For the same instruction, history and current panoramas, a deterministic LLM-based agent may confidently choose a suboptimal perspective (e.g., option B) and make an incorrect action decision. In contrast, humans typically explore several possible futures in their mind, and then question and verify these hypothetical choices before acting. This shows that for VLN, it is not enough to ask an LLM to output a single best action; instead, we should first generate diverse candidate actions and then verify them according to instruction and scenarios before execution.

Motivated by this observation, we propose DV-VLN, a new VLN framework that combines structured navigational chain-of-thought with dual verification at inference time. At each timestep, the backbone LLM receives the instruction, navigation history and textified observations, and produces a small set of candidate actions via sampling decoding, together with their structured reasoning in three fields: Prediction (what the next scene should be), View Match (which view supports this prediction), and Action (which direction to move). DV-VLN then regards each candidate option as a hypothesis and evaluates it through two complementary verification channels: True–False Verification (TFV), which asks whether the candidate option is the correct next step in a given context, and Masked-Entity Verification (MEV), which masks key entities in the instruction and tests whether they can be recovered assuming the candidate option is executed. The final action is determined by adding up TFV and MEV successes over multiple verification samples,  thus generating an interpretable score for each candidate option.

To make this reasoning process reliable and efficient, we adopt a parameter-efficient in-domain training scheme for the LLaMA-2 backbone~\cite{touvron2023llama,touvron2023llama2}. Using the existing VLN trajectory, we build formalized step-wise labels for the navigational chain-of-thought and finetune the LLM so that it can generate Prediction–View Match–Action triples in a unified format. This training method adapts the model to VLN with modest computational cost, while the subsequent dual verification is carried out completely during reasoning and does not require extra supervision. In general, DV-VLN follows a generate–then–verify paradigm: first generate multiple CoT-conditioned actions, then verify and re-rank them before execution, as shown in Fig.~\ref{fig:motivation}.

We conduct extensive experiments on three benchmarks—R2R, RxR (English subset) and REVERIE—and compare DV-VLN with cross-modal and language-only baselines. Under the language-only setting, DV-VLN
achieves competitive results and makes continuous improvements over widely used prompting-based baselines, especially on unseen splits. Despite using no cross-modal pretraining, DV-VLN also achieves competitive performance with several powerful cross-modal backbones, which highlights the effectiveness of verification-guided reasoning. The ablation studies further show that both TFV and MEV bring complementary gains, and that dual verification significantly improves navigation success over naive sampling and voting.

To summarize, our main contributions are as follows: 
\begin{itemize}
\item{We introduce DV-VLN, a first generate-then-verify setup for VLN. Instead of making a single shot decision, it samples several CoT-guided navigation options first, then uses dual verification to re-rank them—making the agent more robust and the reasoning easier to follow.}
\item{We formalize navigation reasoning into Prediction–View Match–Action triples and adapt an open-source LLaMA-2 backbone to this format using parameter-efficient in-domain training on VLN data.}
\item{We design TFV and MEV to explicitly check the consistency of candidate actions with instructions, history and observations, and use their combined verification scores to select actions.}
\item{DV-VLN delivers strong performance among language-only VLN agents and competitive results against cross-modal methods on R2R, RxR and REVERIE, demonstrating the practicality and effectiveness of dual verification for embodied navigation.}
\end{itemize}

\section{Related Work}

\subsection{Vision-Language Navigation}
Vision-Language Navigation (VLN) has attracted extensive attention in recent years, and a rich line of work has been developed around the Room-to-Room and its follow-up benchmarks. Early VLN work had to deal with two key challenges: not enough data, and agents that were hard to train stably. A common approach was data augmentation—back-translation, environment dropout, random environment mixup, or counterfactual path sampling~\cite{tan2019learning, fried2018speaker,liu2021vision,Fu2019CounterfactualVN} People also strengthened the learning setup itself, for example with contrastive instruction–trajectory learning, extra self-supervised reasoning objectives, and adversarial training to improve robustness~\cite{liang2022contrastive,zhu2020vision,lin2021adversarial,lin2022adapt}. Meanwhile, a series of works focused on model architectures, e.g., reinforced cross-modal matching, self-monitoring agents, graph-based global planners, and object-aware modules~\cite{wang2019reinforced,ma2019self, deng2020evolving,qi2020Object}, which improved the cross-modal fusion of language and vision and led to steady gains on standard VLN metrics.

To further enhance generalization to unseen environments, pretraining-based cross-modal backbones became a dominant trend. Representative models~\cite{hong2021vln,Chen2021HistoryAM,Chen2022ThinkGA,Qiao2022HOPHA,Guhur2021AirbertIP,an2022bevbert,wang2023scaling} utilize large-scale vision–language data or in-domain pretraining to learn stronger representations before finetuning on downstream navigation tasks. These methods achieve state-of-the-art results on R2R, RxR and REVERIE, but they often require powerful visual encoders and a large amount of cross-modal pretraining, which limits their scalability and deployment capabilities in resource-constrained scenarios.

More recently, several works have begun to explore LLM-driven VLN agents that operate mainly in the text space. By recasting navigation observations and histories into a textual format, NavGPT \cite{zhou2023navgpt} explores the potential of GPT-style models as zero-shot navigators. This approach effectively shifts the focus from task-specific training to the inherent reasoning capabilities of LLMs. DiscussNav~\cite{long2023discuss} uses multiple expert LLMs and encourages the agent to “discuss before moving”, while MapGPT~\cite{chen2024mapgpt} builds an online language-formalized map to support global exploration and adaptive planning. Affordance-oriented planning with foundation models~\cite{chen2024affordances} further emphasizes leveraging general world knowledge for continuous navigation. NaviLLM~\cite{zheng2024towards} moves towards a trainable generalist navigation model based on LLMs, and NavCoT~\cite{lin2025navcot} introduces a trainable navigational chain-of-thought that decomposes VLN reasoning into structured intermediate steps to better adapt open-source LLMs to navigation tasks. A significant bottleneck for these agents is their heavy reliance on large, closed-source models. Moreover, by primarily focusing on a direct mapping from inputs to actions, they fail to establish a transparent loop for verifying or refining their internal reasoning.

In this work, we follow the language-only line but introduce a new VLN framework, DV-VLN, which combines a trainable open-source LLM backbone with an explicit navigational chain-of-thought and a dual verification mechanism. To enhance the dependability of action selection, DV-VLN adopts a generate–then–verify reasoning strategy. This approach, anchored in a Prediction–View Match–Action structure, utilizes True–False and masked-entity re-ranking to deliver the interpretability of LLM-based agents without requiring extensive cross-modal pretraining.

\subsection{LLMs for Embodied AI}
The impressive planning and reasoning prowess of LLMs has catalyzed their adoption as high-level controllers in embodied AI~\cite{Ahn2022DoAI,Huang2022InnerME,Yao2022ReActSR,shahlm,schumann-2023-velma,wang2023voyager}. SayCan~\cite{Ahn2022DoAI} combines an LLM with affordance functions to generate feasible household plans, while Inner Monologue~\cite{Huang2022InnerME} injects environment feedback into the language loop for iterative refinement. ReAct-style methods~\cite{Yao2022ReActSR} interleave reasoning and acting, and LM-Nav~\cite{shahlm}, VELMA~\cite{schumann-2023-velma} and Voyager~\cite{wang2023voyager} adapt LLMs to navigation and open-ended exploration by grounding language in visual observations.

Recent works further explore in-domain adaptation of LLMs for embodied control. EmbodiedGPT~\cite{mu2023embodiedgpt} constructs a large-scale embodied planning dataset and introduces an additional embodied transformer, while Octopus~\cite{yang2023octopus} and RT-2~\cite{brohan2023rt} integrate environment feedback or web-scale vision–language data to train vision-language-action models. Such efforts suggest that pure prompting reaches a bottleneck; specialized supervision is thus pivotal in aligning internet-scale knowledge with the demands of embodied decision-making.

Positioned as a training-efficient alternative for indoor VLN, DV-VLN shifts away from the reliance on massive robotic datasets or complex action heads. Instead, we leverage an open-source LLaMA-2 backbone, fine-tuned on navigational chain-of-thought labels via parameter-efficient techniques. To safeguard its decision-making, we fortify this lightweight architecture with a dual verification module during the inference phase.

\subsection{Structured Chain-of-Thought Reasoning}
By eliciting intermediate reasoning paths, Chain-of-Thought (CoT) prompting ~\cite{wei2022chain} enables LLMs to tackle complex problems instead of jumping straight to the final output. Later studies enhance CoT through self-consistency, which samples and aggregates multiple reasoning paths~\cite{wang2022self},  least-to-most prompting that solves problems via ordered subproblem decomposition~\cite{zhou2022least}, and bootstrapped reasoning strategies~\cite{zelikman2022star}. By organizing reasoning into tree-based search processes, tree-of-thought methods~\cite{yao2023tree, long2023large} facilitate a more strategic exploration of the thought space prior to the final execution.

Nevertheless, most current methods employ CoT in an open-ended, offline manner. Their reasoning is typically unstructured, without the benefit of task-specific signals to guide the intermediate logical flow. This gap hinders their direct deployment in embodied tasks, which demand rigid reasoning formats and precise visual grounding. DV-VLN addresses this by employing a structured, trainable CoT where navigation logic is decomposed into Prediction, View Match, and Action, all supervised in a step-wise manner. This unified CoT format not only yields interpretable navigation but also serves as the basis for our dual verification module, which treats each CoT-conditioned action as a hypothesis to be explicitly checked before execution.

\subsection{Verification-Based Reasoning and Self-Verification}
A complementary line of work to chain-of-thought prompting studies answer verification. Earlier efforts usually introduce an extra scoring model to filter candidates. This reliance on external modules will increase the complexity of the model due to additional costs, usually at the expense of intuitive interpretability.

Self-Verification~\cite{weng2023large}  proposes a simple yet powerful alternative solution: generate–then–verify without the need for additional training. For a given question,  the model first performs forward CoT reasoning with sampling to obtain multiple candidate answers. Subsequently, each candidate solution will be reviewed by the same large language model, and during the review process, carefully designed verification prompts will be used. Two key mechanisms have been introduced: Truthfulness Verification (TFV), where the candidate solution is rephrased as a declarative statement and the model is asked to determine whether it is true or false; and Conditional Masking Verification (CMV), where part of the problem description is masked and the question is asked whether the masked content can be correctly reconstructed under the assumption that the candidate solution is correct. Verification scores are computed by counting successful checks over multiple samples, and the candidate with the highest score is selected. This framework has seen an improvement in its performance in mathematics, logical reasoning, and question-answering tasks. Moreover, it can operate without training and has a high degree of interpretability.

Inspired by this generate–then–verify paradigm, DV-VLN extends its application to the embodied navigation field. Rather than scrutinizing standalone textual outputs, we validate CoT-conditioned actions within a complex interplay of instructions, temporal history, and real-time visual observations.  We design two navigation-specific verification channels—True–False Verification (TFV) and Masked-Entity Verification (MEV)—that mirror the ideas of judgment-style and masked-condition checks, respectively. TFV asks whether a candidate action is the correct next step given the navigational CoT, while MEV masks key instruction entities (e.g., “bathroom”, “sofa”) and tests whether they can be recovered when the candidate is assumed to be executed. By aggregating success counts from TFV and MEV to re-rank actions, DV-VLN introduces the advantages of self-verification—being training-free, interpretable, and model-agnostic—directly into the VLN domain.

\section{Preliminaries}
\label{Preliminaries}
\subsection{Problem Setup}
Vision-Language Navigation (VLN) tasks an agent with following a natural-language instruction $I$ to navigate from a start to a goal viewpoint. At step $t$, the agent receives a panoramic observation $O_t$, which consists of $K$ single-view images $O_{t,k}$, each noted with its heading and elevation $(\psi_{t,k}, \theta_{t,k})$:
\begin{equation}
O_t = \{O_{t,k}\}_{k=1}^{K}.
\end{equation}
The action space comprises $N$ navigable options among the $K$ views, plus a $[stop]$ token. The agent selects an action $a_t \in \{1, \dots, N, \text{[stop]}\}$ based on the current observation and its navigation history $H_t = \{a_0, \dots, a_{t-1}\}$. A trajectory is successful if the final stop is within 3 m of the goal. Performance is reported via standard navigation metrics (SR/SPL/NE/TL); for RxR, we additionally include nDTW and sDTW to provide a more comprehensive assessment.

\subsection{Large Language Models (LLMs)}
To ensure flexible and controllable language understanding and decision-making for our embodied navigation task, we adopt an open-source Large Language Model (LLM) that can be trained and deployed locally as our core navigation language backbone. Specifically, we utilized the LLaMA-2 series of models released by Meta AI.

We select the LLaMA-2-7B variant, which strikes a balance between computational efficiency and performance for deployment. For efficient in-domain adaptation to the specific characteristics of the navigation environment and instructions, we employ a Parameter-Efficient Bias Tuning strategy. This method enables us to effectively fine-tune the model by merely updating a small portion of the parameters, thereby reducing the huge computational and storage costs associated with conducting a comprehensive fine-tuning of large models.

\begin{figure*}[t]
\begin{centering}
\includegraphics[width=0.98\linewidth]{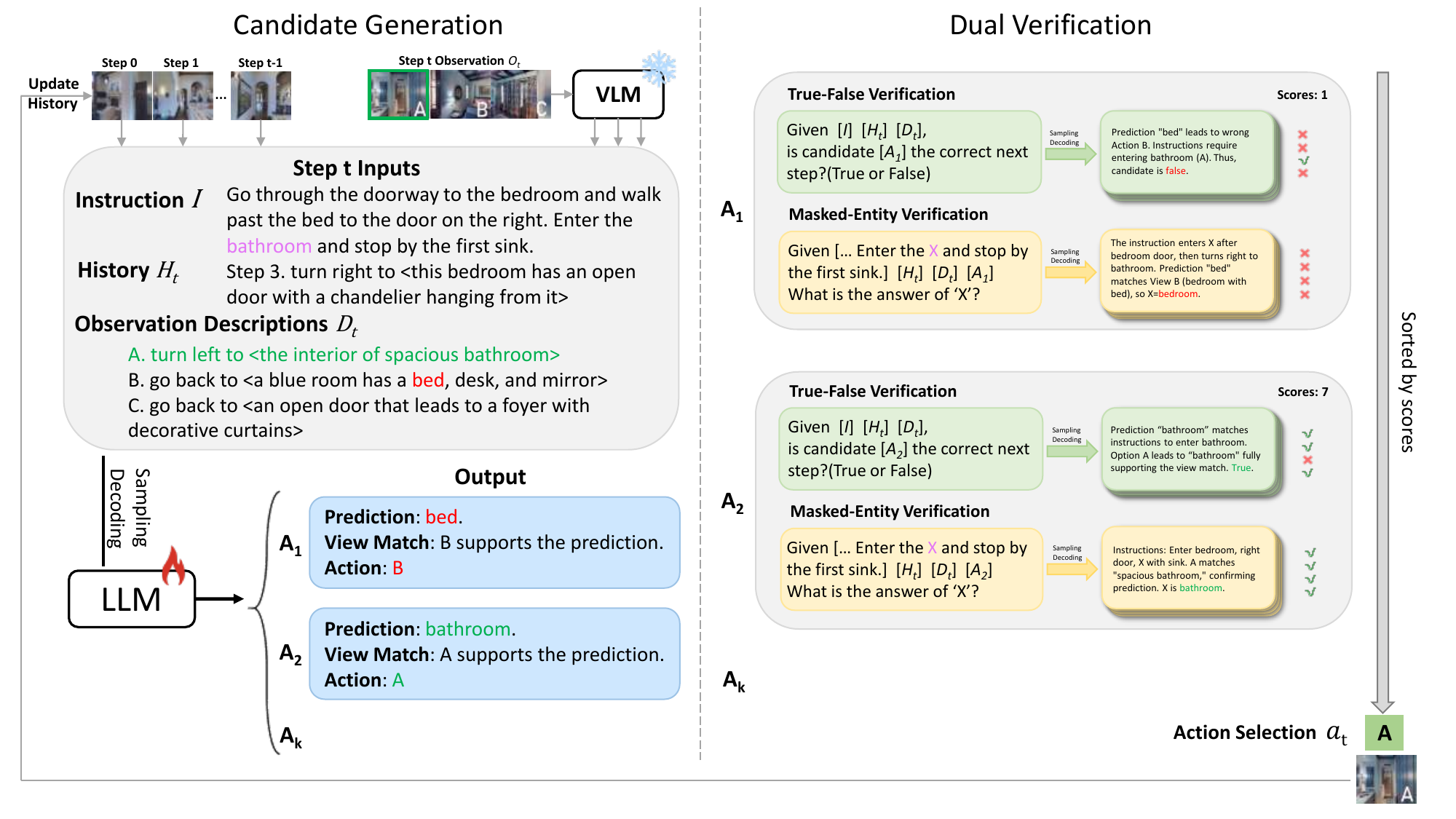}
\par\end{centering}
\caption{Overview of DV-VLN. At timestep $t$, a vision-to-text module converts the panoramic
observation $O_t$ into textual observation descriptions $D_t$, which, together with the instruction $I$ and
navigation history $H_t$, form the navigation input. The LLM performs sampling decoding to generate
$K$ candidate outputs $\{A_k\}$ in a structured navigational chain-of-thought format (Prediction, View
Match, Action). DV-VLN then applies \textbf{dual verification}—True-False Verification (TFV) and Masked-
Entity Verification (MEV)—to score each candidate via multiple verification samples, and selects the
final action $a_t$ by ranking candidates according to their summed verification scores.
}\label{fig:overview}
\vspace{-0.4cm}
\end{figure*}

\section{Method}
We propose DV-VLN, an end-to-end VLN framework that maintains the navigation backbone intact while equipping it with an inference-time Dual Verification module. As illustrated in Fig.~\ref{fig:overview}, at each timestep $t$ DV-VLN:
\begin{enumerate}
\item converts the panoramic view into textual observations via a vision-to-text interface (Sec.~\ref{Vision-to-Text-System});
\item prompts the LLM to produce a three-step navigational chain-of-thought (CoT) consisting of Prediction, View match and Action (Sec.~\ref{Design of Chain-of-Thought Prompt});
\item collects step-wise CoT ground truth from VLN data and trains the backbone with multi-task pretraining and imitation-style finetuning (Sec.~\ref{Ground-Truth Collection}--~\ref{Training and Inference});
\item at inference, generates multiple candidate actions and applies Dual Verification (TFV + MEV) to rescore and re-rank them by counting verification successes (Sec.~\ref{Dual Verification at Inference}).
\end{enumerate}

\subsection{Textual Observation Construction}
\label{Vision-to-Text-System}
At timestep $t$, the agent receives the panoramic observation $O_t = \{O_{t,n}\}_{n=1}^K$ where each view $O_{t,n}$ consists of an RGB image $B_{t,n}$ and direction information $A_{t,n} = \{\psi_{t,n}, \theta_{t,n}\}$ (heading and elevation). As shown on the left side of Fig.~\ref{fig:overview}, DV-VLN first translates each view into a compact textual description before passing it to the LLM:

\noindent \textbf{Image captioning.} We use the image captioning model BLIP (denoted $F_v$) to convert each RGB image into a visual caption:
\begin{equation}
    D_{t,n}^v = F_v(B_{t,n}).
\end{equation}

\noindent \textbf{Direction mapping.} The direction $A_{t,n}$ is mapped into a discrete textual direction space containing six basic directions (e.g., ``turn left'', ``turn right'', ``go forward'', ``go back'', ``go up'', ``go down'') following standard VLN direction mapping rules. We denote the mapped phrase by $D_{t,n}^a$.

\noindent \textbf{Final textual description.} We concatenate the direction phrase and visual caption to obtain the textual description of each navigable view:
\begin{equation}
   D_{t,n} = \text{cat}(D_{t,n}^a, D_{t,n}^v), 
\end{equation}

where $\text{cat}(\cdot, \cdot)$ denotes string concatenation.

Collecting all $N$ navigable views yields the textual observation set $D_t = \{D_{t,n}\}_{n=1}^N$. For convenience, each view is assigned an alphabetical label (\textbf{A, B, C}, ...), which will serve as action options in the subsequent reasoning and verification stages. The LLM then takes ($I_t, H_t, D_t$) as input at each timestep.

\subsection{Navigational Chain-of-Thought Prompt}
\label{Design of Chain-of-Thought Prompt}

Designing informative \textbf{intermediate reasoning steps} is crucial for chain-of-thought prompting: the structure of the CoT strongly influences the quality and stability of the final action prediction.
Inspired by the idea that an agent should both anticipate future observations and ground them in the current scene, we formalize a \textbf{three-step navigational chain-of-thought} tailored to VLN:

\noindent\textbf{Prediction (Pred).} A short hypothesis about the next object or scene the agent expects to see.

\noindent\textbf{View match (VM).} An explicit decision about which option best supports the prediction.

\noindent\textbf{Action (Act).} The final navigation action option to execute.
We denote the navigational chain-of-thought at timestep $t$ as:
\begin{equation*}
    \text{CoT}_t = (\text{Pred}_t, \text{VM}_t, \text{Act}_t).
\end{equation*}

As shown in the middle block of Fig.~\ref{fig:overview}, at each timestep $t$ the LLM receives the instruction $I$, the textually described observation $D_{t}$ obtained in Sec.~\ref{Vision-to-Text-System}, i.e., $D_{t}$ = $\{D_{t,n}\}^{N}_{n=1}$, and the navigation history $H_{t}$. In addition, we provide the LLM with one in-context CoT example that demonstrates the desired reasoning format and principles. This example teaches the model how to systematically relate instruction entities, visual observations, and previous actions.

Given $(I, H_t, D_t)$ and the example, the LLM is asked to output the three-step CoT in a constrained textual format:

\noindent\textbf{Prediction (Pred).}
We ask the model to anticipate what it will encounter next, conditioned on the instruction and history. The desired format is: 

\texttt{Prediction: $U_{t}$}.

where $U_t$ is a short phrase such as ``bathroom'', ``open door'', or ``table near the window''. This step captures the agent's internal expectation of upcoming landmarks and provides a semantic target for subsequent grounding.

\noindent\textbf{View match (VM).}  
After predicting $U_t$, the model must select which candidate view in $D_t$ best matches this prediction. Denote the chosen option letter by $V_t$ (e.g., \textbf{A}, \textbf{B}, \textbf{C}). The desired format is:

\texttt{View match: $V_{t}$ supports the prediction.}

\noindent\textbf{Action (Act).} 
Finally, the model summarizes its reasoning and proposes the next action option $a_t$ (which can be the same as $V_t$ or differ, depending on the navigation dynamics). The format is:

\texttt{Action: $a_{t}$.}

We give an in-context example to the LLM as follows:

\texttt{Input: Instruction: Walk past the sofa and stop at the bathroom door. Observation: [A. stop,  B. go forward to <a sofa>, C. turn right to <a bathroom door>]. History: Step 1. go forward to <a sofa>.}

\texttt{Output: Prediction: bathroom door. View match: C supports the prediction. Action: C.}

This example shows that the model first tracks progress (sofa already reached), then predicts what comes next (bathroom door), grounds this prediction in option C, and finally chooses C as the action. Based on such in-context examples, we prompt the LLM at each timestep with:

\texttt{Input:~Instruction:~\{$I$\} Observation: \{$D_{t}$\} History:~\{$H_{t}$\}}


\texttt{Output:}

\subsection{Navigational CoT Ground-Truth Collection} \label{Ground-Truth Collection}
Although large language models can produce plausible multi-step reasoning, in complex VLN scenes their zero-shot CoT is often noisy or inconsistent, which may lead to wrong actions. To enable \textbf{reliable in-domain training} of the navigational CoT, we construct step-wise ground-truth CoT labels $\text{CoT}_t^*$ from existing VLN data.

\noindent\textbf{Ground-truth Prediction label.} We first build the ground-truth label $U_t^*$ for the \textbf{Prediction} step. Ideally, $U_t^*$ should satisfy two properties:
\begin{enumerate}
    \item It should correspond to an object/scene that actually appears in the \textbf{next ground-truth observation} $B_t^*$ .
    \item It should be \textbf{mentioned in the instruction $I$}, to encourage strong alignment between language and perception.
\end{enumerate}
To achieve this, we proceed in two stages:

\noindent\textbf{Instruction entity extraction.} We prompt an LLM to extract all landmarks, objects, and scene phrases mentioned in $I$, obtaining a list
    \begin{equation*}
    U^{\text{la}} = \{U_k^{\text{la}}\}_{k=1}^M,
    \end{equation*}
    where $U_k^{\text{la}}$ is a candidate entity (e.g., ``sofa'', ``bathroom door'', ``window''), and $M$ is the number of extracted entities.
    
\noindent\textbf{Vision-language matching.} For each timestep $t$, let $B_t^*$ denote the ground-truth next observation image. We compute the similarity between $B_t^*$ and each entity $U_k^{\text{la}}$ using a vision-language model such as CLIP with text encoder $F_{\text{CLIP}}^{\text{text}}$ and image encoder $F_{\text{CLIP}}^{\text{img}}$, and select the highest-scoring entity as the Prediction label:
    \begin{equation}
    U_t^* = \underset{U_k^{\text{la}}}{\arg \max} \operatorname{Sim} \left( F_{\text{CLIP}}^{\text{text}}(U_k^{\text{la}}), F_{\text{CLIP}}^{\text{img}}(B_t^*) \right).
    \end{equation}
This procedure yields a Prediction label that both appears in the instruction and is visually supported by the ground-truth next view.

Since the reasoning task VM aims to find aligned observation with the instruction for action decision, we set the ground-truth label of View match to be consistent with the option of ground-truth action $a^*_t$. Combining the three components, the ground truth of the navigational chain-of-thought, denoted as $\mathrm{CoT}^{*}_{t}$, for instruction $I$ at timestep $t$ is defined by:

\texttt{Prediction:~${U}^{*}_{t}$.~View~match: $a^{*}_{t}$ matches the imagination. Action:~$a^{*}_{t}$.}

These CoT labels serve as supervision for in-domain training in Sec.~\ref{Training and Inference}, enabling the LLM to learn structured, disentangled navigation reasoning that is tightly coupled with both the instruction and the visual trajectory.

\subsection{In-domain Navigational CoT Training}
Pretraining and finetuning are standard tools for training VLN agents. In DV-VLN, we adopt an in-domain training scheme that explicitly leverages the \textbf{navigational CoT structure} defined in Sec.~\ref{Design of Chain-of-Thought Prompt} and the ground-truth labels $\mathrm{CoT}^{*}$ from Sec.~\ref{Ground-Truth Collection}. The goal is to make the LLaMA-2 backbone reliably generate high-quality CoTs that follow the desired format and support accurate action decisions.

\label{Training and Inference}


\noindent\textbf{Pretraining.}
We first decompose the three CoT steps---Prediction, View match, and Action---into separate supervised tasks, and perform \textbf{multi-task pretraining}. This encourages each reasoning component to be accurate on its own, which in turn stabilizes the full CoT. 

Let $D$ denote the textual observations for step $t$ and $H$ the action history. We denote the model outputs for the three steps by $U$, $V$, and $a$, and the corresponding ground-truth labels by $U^*$, $V^*$, and $a^*$ as obtained in Sec.~\ref{Ground-Truth Collection}. The pretraining objectives for each task are:
\begin{equation}
\mathcal{L}_{\mathrm{Pred}} = -U^{*}\mathrm{log}(p_{\mathrm{LLM}}(U|I,H,D)),
\end{equation}
\begin{equation}
\mathcal{L}_{\mathrm{VM}} = -V^{*}\mathrm{log}(p_{\mathrm{LLM}}(V|I,H,D)),
\end{equation}
\begin{equation}
\mathcal{L}_{\mathrm{Act}} = -a^{*}\mathrm{log}(p_{\mathrm{LLM}}(a|I,H,D)),
\end{equation}

The total pretraining loss is the sum:
\begin{equation}
\mathcal{L}_{p}=\mathcal{L}_{\mathrm{Pred}}+\mathcal{L}_{\mathrm{VM}}+\mathcal{L}_{\mathrm{Act}},
\end{equation}
This stage teaches the model, for each individual step, to (i) predict an instruction-consistent landmark, (ii) select the aligned view, and (iii) choose the correct action option.

\noindent\textbf{Finetuning.}
To create a dataset for instruction following, we convert the expert trajectories from the original VLN dataset utilizing $\text{CoT}^*_t$. At every step $t$, the model is trained to produce a full navigational chain-of-thought $\text{CoT}_t$ for action decisions. The finetuning objective $\mathcal{L}_f$ is defined as:
\begin{equation}
\mathcal{L}_{f} = -\sum_{t}\mathrm{CoT}^{*}_{t}\mathrm{log}(p_{\mathrm{LLM}}(\mathrm{CoT_{t}}|I,H_{t},D_{t})).
\end{equation}

Following the completion of in-domain training, we instruct the models to formulate navigational chains-of-thought to facilitate action selection, utilizing the prompts and in-context exemplars detailed in Sec.~\ref{Design of Chain-of-Thought Prompt}. By leveraging our proposed training methodology, the LLM acquires the ability to synthesize structured navigational reasoning in the requisite format, thereby enhancing the accuracy of its self-guided decision-making processes.

\subsection{Dual Verification at Inference}
\label{Dual Verification at Inference}
As illustrated in Fig.~\ref{fig:overview} (right), DV-VLN performs \textbf{dual posterior verification} on the action candidates produced by sampling decoding. At each timestep $t$, the backbone agent yields $K$ candidate actions $\{a_t^k\}_{k=1}^K$ together with their navigational CoT outputs (Prediction--View Match--Action). DV-VLN does not alter this front-end reasoning; instead, it evaluates each candidate through two complementary backward checks, then re-ranks candidates by a simple additive verification score. This design follows the self-verification principle that ``generate first, then verify by repeated sampling'' yields interpretable confidence signals without training extra verifiers.

\noindent\textbf{True--False Verification (TFV).} For each candidate $a_t^k$, we construct a judgement prompt that asks whether the candidate is a correct next step under the given instruction $I$, history $H_t$, and current observation descriptions $D_t$ (see Fig.~\ref{fig:overview}, green block). We query the verifier LLM with sampling decoding $P$ times to obtain boolean outputs $\{b_{k,p}\}_{p=1}^P$, where $b_{k,p} \in \{\text{True}, \text{False}\}$. The TFV score is the number of times the verifier returns \textit{True}.
\begin{equation}
    S_{\text{TFV}}(k) = \sum_{p=1}^{P} \mathbf{1}(b_{k,p} = \text{True}).
\end{equation}
\noindent\textbf{Masked-Entity Verification (MEV).} TFV checks global action correctness, while MEV probes whether the candidate implicitly preserves the semantics of key instruction entities (Fig.~\ref{fig:overview}, yellow block). We select $R$ salient entities/scenes in $I$ (e.g., rooms or landmarks), mask one entity at a time to form $\{I_t^r\}_{r=1}^R$, and ask the verifier to recover the masked content given $(I_t^r, H_t, D_t, a_t^k)$. For each masked instruction $r$, we sample $P$ verifier outputs $\{\hat{e}_{k,p}^r\}_{p=1}^P$. Let $e_t^r$ denote the original  entity. The MEV score counts how often the verifier correctly reconstructs the entity.

\begin{equation}
    S_{\text{MEV}}(k) = \sum_{r=1}^{R} \sum_{p=1}^{P} \mathbf{1}(\hat{e}_{k,p}^r = e_t^r).
\end{equation}
\noindent\textbf{Additive verification score and action selection.} DV-VLN uses a simple, training-free fusion: the final verification score of candidate $k$ is the sum of the two posterior checks. This yields an interpretable ``how many times it passes verification'' measure.
\begin{align}
S(k) &= S_{\text{TFV}}(k) + S_{\text{MEV}}(k), \\
a_t &= \arg \max_{k \in \{1, \dots, K\}} S(k).
\end{align}
When multiple candidates tie, we choose the one with higher TFV score; if still tied, we fall back to the earliest candidate in decoding order. In practice, if the $K$ sampled candidates are already consistent (all predict the same action), DV-VLN directly executes it and skips verification to save computation; otherwise the dual verification branch is activated (Fig.~\ref{fig:overview}, right).

\begin{table*}[t]

	
	\resizebox{1.0\linewidth}{!}{
	{\renewcommand{\arraystretch}{1.1}
		\begin{tabular}{ccc||c|c|c|c|c|c|c|c|c|c}

			\specialrule{.1em}{.05em}{.05em}
			
				\multirow{2}{*}{Setting}&\multirow{2}{*}{Method}&\multirow{2}{*}{Year}&\multicolumn{5}{c|}{Val Seen}&\multicolumn{5}{c}{Val Unseen}\cr\cline{4-13}
            
			&&&TL&NE $\downarrow$&OSR $\uparrow$&SR $\uparrow$&SPL $\uparrow$&TL&NE $\downarrow$&OSR $\uparrow$&SR $\uparrow$&SPL $\uparrow$\cr
			\hline
			
            \multirow{6}{*}{\makecell{Cross-modal \\backbone}}&Seq2Seq~\cite{anderson2018vision}& 2018 &11.33&6.01&53&39&-&8.39&7.81&28&21&-\\
            &Speaker Follower~\cite{fried2018speaker}& 2018 &-&3.36&74&66&-&-&6.62&45&36&-\\
            &HAMT~\cite{Chen2021HistoryAM}& 2021 &11.15&2.52&-&76&72&11.46&2.29&-&66&61\\
            &DUET~\cite{Chen2022ThinkGA}& 2022 &12.32&2.28&86&79&73&13.94&3.31&81&72&60\\
            &BEVBert~\cite{an2022bevbert}& 2022 &13.56&2.17&88&81&74&14.55&2.81&84&75&64\\
            &ScaleVLN$^{\ddagger}$~\cite{wang2023scaling}& 2023 &13.24&\textbf{2.12}&\textbf{87}&\textbf{81}&\textbf{75}&14.09&\textbf{2.09}&\textbf{88}&\textbf{81}&\textbf{70}\\
            \hline
            \multirow{5}{*}{\makecell{Prompting-based \\LLM agents}}&NavGPT~\cite{zhou2023navgpt}& 2023 &-&-&-&-&-&11.45&6.46&42&34&29\\
            &DiscussNav~\cite{long2023discuss}& 2023 &-&-&-&-&-&9.69&5.32&61&43&40\\
            &MapGPT~\cite{chen2024mapgpt}& 2024 &-&-&-&-&-&-&5.63&58&44&35\\
            &MSNav~\cite{liu2025msnav}& 2025 &-&-&-&-&-&-&5.24&65&46&40\\
            &NavGPT-2$^{\dag}$~\cite{zhou2024navgpt}& 2024 &14.13&2.84&83&74&63&14.01&2.98&84&74&61\\
            \hline
            \multirow{4}{*}{\makecell{Trainable \\language-only agents}}&NaviLLM$^{\ddagger}$~\cite{zheng2024towards}& 2024 &-&-&-&-&-&12.81&3.51&-&67&59\\
            &EvolveNav*~\cite{lin2025evolvenav}& 2025 &-&-&-&-&-&12.07&3.15&-&71&63\\
            &NavCoT~\cite{lin2025navcot}& 2025 &10.08&6.46&48&41&38&9.95&6.26&48&40&37\\
            &DV-VLN (ours)& 2025 &10.53&5.03&66&54&49&10.70&5.14&67&52&45\\            
    
 \specialrule{.1em}{.05em}{.05em}

		\end{tabular}}}
  \vspace{-0.2cm}
  \caption{Comparison with SOTAs on R2R. $^{\dag}$ denotes using additional synthetic trajectories from PREVALENT for training. $^{\ddagger}$ denotes multi-task pretraining on multiple embodied datasets beyond R2R. * denotes iterative self-training with self-enriched chain-of-thought supervision.}
	\label{tab:com to sota}
	\vspace{-0.2cm}
\end{table*}

\section{Experiments}
\subsection{Experimental Setup}
\noindent\textbf{Datasets.} 
We conduct experiments on three widely used VLN benchmarks: R2R (Room-to-Room), RxR (English subset), and REVERIE. R2R is built on 90 real indoor scans and contains 7,189 navigation trajectories, each paired with three fine-grained instructions. RxR offers longer routes and substantially more complex instructions than R2R. Since our masked-entity verification relies on CLIP-style text–vision alignment which is pretrained on English data, we follow common practice and evaluate on the English split of RxR, including 26,464 training, 2,939 Val Seen, and 4,551 Val Unseen instruction–path pairs. REVERIE replaces R2R’s fine-grained directives with higher-level, goal-oriented instructions, serving as a strong test of cross-dataset generalization.

\noindent\textbf{Evaluation Metrics.}
For R2R and REVERIE, we report the standard VLN metrics: 1) Trajectory Length (TL): the average length of the predicted path; 2) Navigation Error (NE): the average distance between the final position of the agent and the target position; 3) Success Rate (SR): the proportion of rounds in which the agent stops within 3 meters of the target;  4) Success rate weighted by Path Length (SPL): the result of dividing SR by the ratio of the shortest path length to the predicted path length; 5) Oracle Success Rate (OSR): success if any visited viewpoint sees the target. For RxR, where instruction-following quality is also emphasized, we additionally report the Coverage weighted
by Length Score (CLS), the normalized Dynamic Time
Warping (nDTW), and the Success weighted by nDTW
(SDTW), which measure path coverage and trajectory–instruction alignment.

\noindent\textbf{Implementation Details.} We finetune LLaMA-2-7B via bias tuning, resulting in about 1.6M trainable parameters. Training is carried out on 4×V100 GPUs with batch size 8, taking roughly 10 hours in total. Inference is run on a single V100 GPU. We optimize with AdamW, using learning rate 0.001 and weight decay 0.02. To accelerate evaluation of ablations, we also adopt a compact Val Unseen Subset for R2R, randomly sampling 90 instruction–trajectory pairs from 8 scans in Val Unseen; this subset provides a fast yet reliable testbed for comparing variants.
\label{Experimental Setup}

\subsection{Comparison on R2R}

Table~\ref{tab:com to sota} reports the results on the R2R benchmark. For a clear and fair comparison, we group prior approaches into (i) fully supervised cross-modal backbones, (ii) prompting-based LLM agents, and (iii) trainable language-only agents.

\noindent\textbf{Cross-modal backbones.} The first block includes fully supervised architectures such as HAMT, DUET, BEVBert, and ScaleVLN, which rely on heavy visual encoders and large-scale vision–language pretraining. These models represent the strongest upper bound on R2R; for example, ScaleVLN achieves 81\% SR and 70\% SPL on Val Unseen. While DV-VLN does not employ cross-modal pretraining or a dedicated visual encoder, it still attains 52\% SR and 45\% SPL on Val Unseen, substantially outperforming early supervised baselines (e.g., Seq2Seq and SpeakerFollower). This indicates that verification-guided decision making can noticeably improve action quality even under a lightweight, language-centric setting.

\noindent\textbf{LLM-based agents.} The second and third blocks summarize LLM-driven agents. Among prompting-based methods (NavGPT, DiscussNav, MapGPT, MSNav), DV-VLN yields clear gains, improving Val Unseen performance to 52\% SR / 45\% SPL compared with the strongest prompting baseline (46\% SR / 40\% SPL of MSNav). This demonstrates the benefit of moving from single-shot prompting to a generate–verify–select inference paradigm. In the trainable language-only category, DV-VLN remains competitive against stronger training recipes such as NavGPT-2 (trained with additional synthetic trajectories), NaviLLM (multi-task pretraining on multiple embodied datasets beyond R2R), and EvolveNav (iterative self-training with self-enriched CoT supervision). Despite using a substantially lighter training and data setting, DV-VLN achieves consistently strong results on both Val Seen and Val Unseen splits, highlighting the effectiveness and portability of dual verification for improving navigation reliability.

\noindent\textbf{Discussion.} Recent LLM-based VLN agents show that language-centric navigation is attractive for its scalability and interpretability: the agent can explicitly articulate intermediate reasoning and decisions, which is difficult for conventional cross-modal transformers. However, performance often depends strongly on the availability of large external resources (e.g., synthetic trajectories, multi-dataset pretraining, or iterative self-training), and may vary across settings. In contrast, DV-VLN focuses on a simple yet effective inference-time principle—dual verification for candidate re-ranking—providing a plug-in reliability boost with transparent decision evidence (TFV/MEV outcomes) and without requiring heavy cross-modal pretraining. These properties make DV-VLN a practical and explainable baseline for language-only navigation under standard training protocols.

Overall, the R2R comparison suggests that DV-VLN provides a computation-efficient alternative to heavy cross-modal systems and stronger high-resource LLM agents: it delivers large improvements over mainstream prompting-based VLN agents while remaining competitive under standard training protocols, making it a strong baseline for scalable language-centric navigation.

\begin{table}
\fontsize{14}{14}\selectfont
    \resizebox{1.0\linewidth}{!}{
    {\renewcommand{\arraystretch}{1.2}
    \begin{tabular}{ccc|cccc}
    \specialrule{.1em}{.05em}{.05em}
        
            \multirow{2}{*}{Sampling} & \multirow{2}{*}{TFV} & \multirow{2}{*}{MEV} & \multicolumn{4}{c}{Val Unseen Subset}\cr\cline{4-7}
            
             & & & NE $\downarrow$&OSR $\uparrow$&SR $\uparrow$&SPL $\uparrow$\cr

            \hline
        
            $\times$ & $\times$ & $\times$ & 6.26 & 48 & 39 & 37 \\
            
             \checkmark & $\times$ & $\times$ & 6.01 & 53 & 40 & 38 \\
             
             \checkmark & \checkmark & $\times$ & 5.34 & 59 & 42 & 40 \\
             
             \checkmark & $\times$ & \checkmark & 5.26 & 61 & 44 & 42 \\
             
             \checkmark & \checkmark & \checkmark & \textbf{5.11} & \textbf{65} & \textbf{52} & \textbf{47} \\

 \specialrule{.1em}{.05em}{.05em}
    \end{tabular}
}}

\vspace{-0.2cm}
\caption{Ablation results of different components on the R2R Val Unseen Subset.}
\label{ablation}
\vspace{-0.4cm}
\end{table}

\subsection{Ablation Study}
To better understand the contribution of each component in DV-VLN, we conduct an ablation study on the R2R Val Unseen Subset introduced in Sec.~\ref{Experimental Setup}. The results are reported in Table~\ref{ablation}. The row “–” corresponds to the backbone navigator with deterministic decoding (no sampling and no verification). “Sampling (Vote)” enables sampling decoding and selects the final action by majority vote over the sampled actions, but still without any verification. The remaining rows progressively incorporate True–False Verification (TFV), Masked-Entity Verification (MEV), and their combination.

Compared with the deterministic baseline, simply turning on sampling and voting brings only marginal gains: NE decreases from 6.26 m to 6.01 m, SR increases slightly from 39\% to 40\%, and SPL from 37\% to 38\%. This indicates that naively aggregating multiple samples is not sufficient to fully exploit the diversity of candidate actions. Adding TFV on top of sampling leads to a clearer improvement: NE is reduced to 5.34 m, OSR rises from 53\% to 59\%, and SR/SPL reach 42\%/40\%, showing that judgment-style verification already helps the agent filter out globally inconsistent candidates. Introducing only MEV yields a similar but slightly stronger effect: NE further drops to 5.26 m, OSR becomes 61\%, and SR/SPL increase to 44\%/42\%, suggesting that entity-level semantic recoverability provides a complementary signal on whether an action is aligned with the instruction landmarks.

The full DV-VLN variant, Sampling + TFV + MEV, achieves the best performance across all metrics on this subset: NE 5.11 m, OSR 65\%, SR 52\%, SPL 47\%. Relative to the backbone without sampling or verification, this corresponds to 1.15 m reduction in NE and +13 / +10 absolute points in SR / SPL, and even compared with the “Sampling (Vote)” baseline it yields large additional gains. These results confirm that TFV and MEV are both effective and mutually reinforcing: TFV captures global consistency, while MEV enforces entity-level alignment, and their combination is crucial for fully realizing the benefits of candidate generation in DV-VLN.

\subsection{Generalization to Other Datasets}

To assess the robustness of DV-VLN beyond R2R, we further evaluate it on the RxR English subset and REVERIE, keeping the training recipe and inference hyperparameters identical to those used on R2R. The results are shown in Table~\ref{tab:rxr} and Table~\ref{tab:reverie}, respectively.

On the RxR English subset (Table~\ref{tab:rxr}), DV-VLN is compared with two strong cross-modal baselines (EnvDrop and HAMT) and two language-only VLN agents (NavCoT and NavGPT-2). As expected, HAMT still achieves the best overall scores thanks to its powerful multimodal pretraining. Nevertheless, DV-VLN attains 29.16\% SR and 25.94\% SPL, clearly outperforming the language-only baselines (e.g., +4.64 SR and +3.36 SPL over NavCoT, and +0.41 SR and +3.58 SPL over NavGPT-2). Moreover, DV-VLN yields higher CLS (48.41), nDTW (41.77) and SDTW (22.17) than NavCoT, indicating that trajectories generated with dual verification not only succeed more often but also follow the instructions more faithfully. These improvements are particularly notable given the longer, more complex instructions in RxR, and suggest that masked-entity verification (MEV) is well suited for datasets where fine-grained instruction grounding is critical.

On REVERIE (Table~\ref{tab:reverie}), which focuses on high-level, goal-directed instructions, DV-VLN again demonstrates strong generalization. Among language-only methods, DV-VLN substantially outperforms both NavCoT and NavGPT, achieving 29.8\% SR and 21.7\% SPL on Val Unseen, compared with NavGPT’s 19.2\% SR and 14.6\% SPL. Remarkably, DV-VLN also surpasses the cross-modal HAMT in terms of SR (29.8\% vs. 23.8\%), while obtaining a comparable SPL (21.7\% vs. 22.4\%) with a shorter trajectory length (9.24 vs. 9.24 TL, i.e., similar path cost). This shows that a language-only agent equipped with dual verification can rival, and in some metrics surpass, heavily pre-trained cross-modal systems on goal-oriented navigation.

Taken together, the results on RxR and REVERIE demonstrate that DV-VLN generalizes well across datasets with very different instruction styles. Without dataset-specific tuning or additional pretraining, the same dual verification mechanism consistently improves success and instruction adherence, underscoring the portability and effectiveness of our approach.

\begin{table}
\fontsize{17}{17}\selectfont
    \resizebox{1.0\linewidth}{!}{
    {\renewcommand{\arraystretch}{1.2}
    \begin{tabular}{l|ccccc}
    \specialrule{.1em}{.05em}{.05em}
			\multirow{2}{*}{Method}&\multicolumn{5}{c}{Val Unseen}\cr\cline{2-6}
&SR $\uparrow$&SPL $\uparrow$&CLS $\uparrow$&nDTW $\uparrow$&SDTW $\uparrow$\cr
			\hline
			
        EnvDrop~\cite{tan2019learning}&38.5&34&54&51&32\\
        HAMT~\cite{Chen2021HistoryAM}&\textbf{38.26}&\textbf{36.23}&\textbf{58.45}&\textbf{53.08}&\textbf{32.81}\\
        \hline
        
        NavCoT~\cite{lin2025navcot}&24.52&22.58&45.06&38.94&19.63\\
        NavGPT-2~\cite{zhou2024navgpt}&28.75&22.36&-&38.50&19.24\\
        DV-VLN (ours)&\textbf{29.16}&\textbf{25.94}&\textbf{48.41}&\textbf{41.77}&\textbf{22.17}\\

 \specialrule{.1em}{.05em}{.05em}
    \end{tabular}
}}

\caption{Comparison results on RxR English subset.}
\vspace{-0.2cm}
\label{tab:rxr}
\end{table}

\begin{table}[t]
\centering
\small
\resizebox{1.0\linewidth}{!}{%
{\renewcommand{\arraystretch}{1.2}%
\begin{tabular}{l|cccc}
\specialrule{.1em}{.05em}{.05em}
\multirow{2}{*}{Method} & \multicolumn{4}{c}{Val Unseen} \\ \cline{2-5}
 & TL & SR $\uparrow$ & OSR $\uparrow$ & SPL $\uparrow$ \\
\hline
Seq2Seq~\cite{anderson2018vision} & - & 4.2 & 9.1 & 2.8\\
RCM~\cite{wang2019reinforced} & 11.98 & 9.3 & 14.2 & 7.0\\
FAST-MATTN~\cite{qi2020reverie} & 45.3 & 14.4 & 28.2 & 7.2\\
HAMT~\cite{Chen2021HistoryAM} & 9.24 & \textbf{23.8} & \textbf{26.4} & \textbf{22.4}\\
\hline
NavCoT~\cite{lin2025navcot} & 12.36 & 9.2 & 14.2 & 7.2\\
NavGPT~\cite{zhou2023navgpt} & - & 19.2 & 28.3 & 14.6\\
DV-VLN (ours) & 9.24 & \textbf{29.8} & \textbf{36.7} & \textbf{21.7}\\
\specialrule{.1em}{.05em}{.05em}
\end{tabular}
}}%
\caption{Comparison results on REVERIE.}
\label{tab:reverie}
\vspace{-0.4cm}
\end{table}
\normalcolor

\begin{figure*}[h]
\centering
\includegraphics[width=0.48\linewidth]{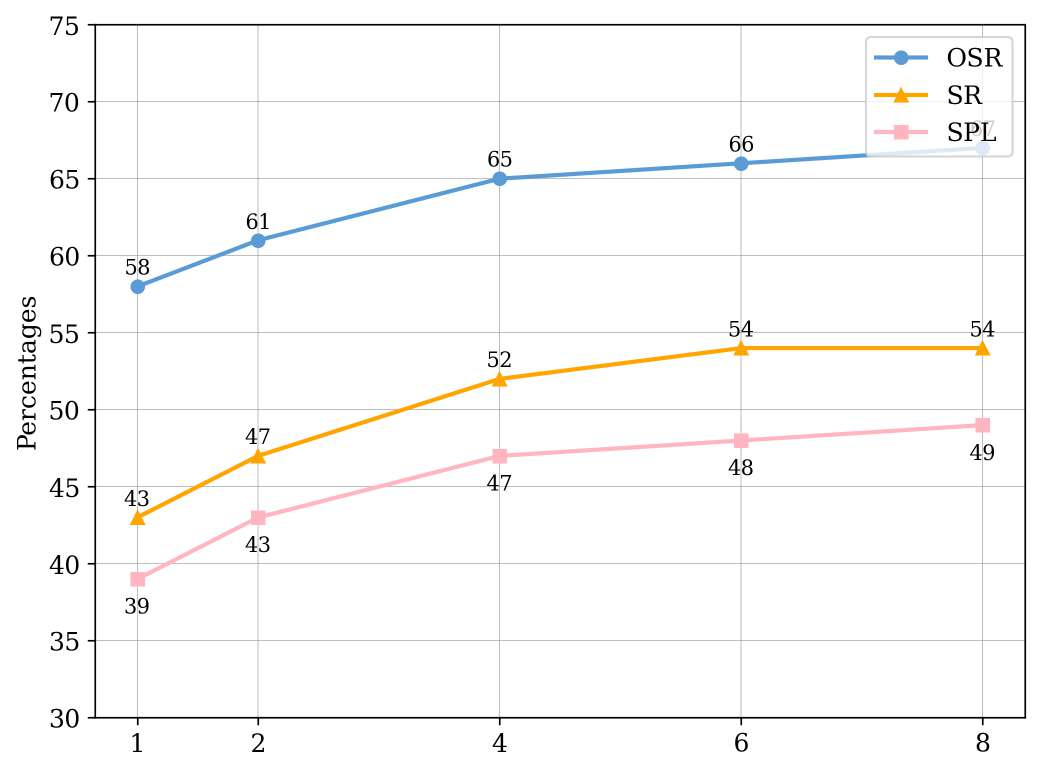}
\hfill  
\includegraphics[width=0.48\linewidth]{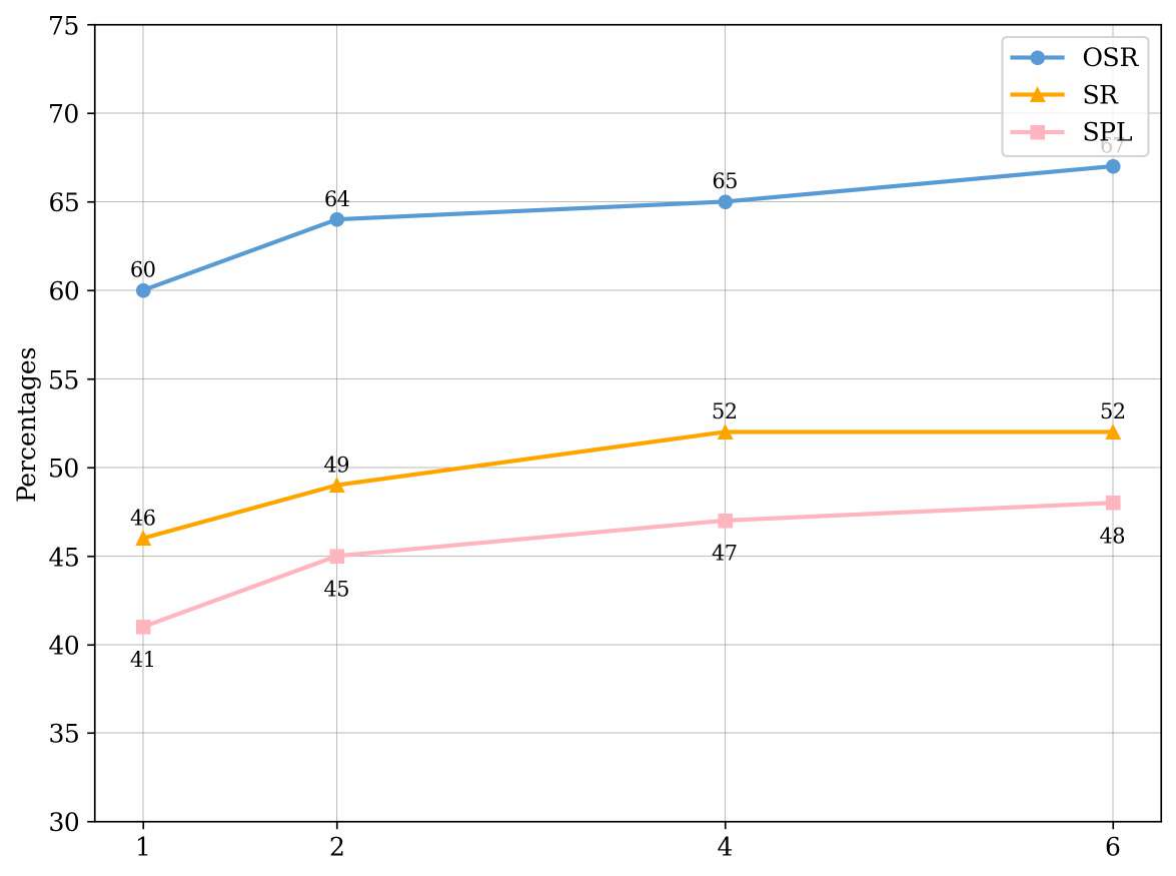}
\caption{Performance sensitivity of DV-VLN on the R2R Val Unseen Subset. Left: varying the number of sampled candidates $K \in \{1,2,4,6,8\}$. Right: varying the number of verification trials $P \in \{1,2,4,6\}$. We report OSR, SR, and SPL.}
\label{fig:kp}
\end{figure*}
\normalcolor

\begin{figure*}[!h]
\begin{centering}
\includegraphics[width=1\linewidth]{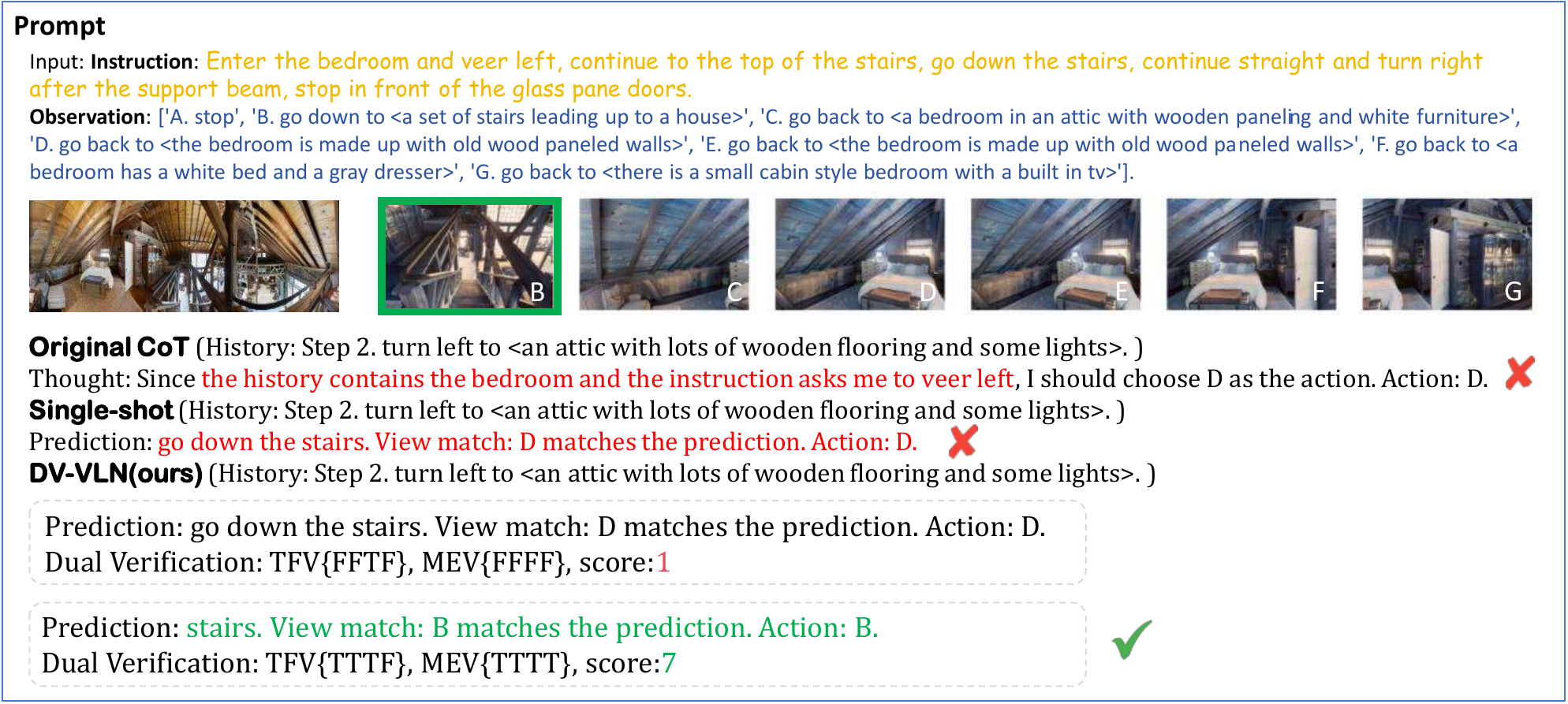}
\par\end{centering}
\caption{Qualitative success case of DV-VLN on R2R. Compared with single-shot decision making, DV-VLN samples multiple CoT-conditioned candidates and re-ranks them via dual verification (TFV+MEV); the correct action (B) receives a higher verification score than the incorrect candidate (D), leading to a reliable final selection.
}\label{case}
\vspace{-0.4cm}
\end{figure*}
\subsection{Performance–Efficiency Trade-off}
DV-VLN improves action reliability by sampling multiple candidates and verifying them multiple times at inference. We therefore study the performance–efficiency trade-off by varying the number of sampled candidates $K$ and the number of verification trials $P$ on the R2R Val Unseen Subset while keeping other settings unchanged. Fig.~\ref{fig:kp} reports OSR, SR, and SPL under different K and P.

\noindent\textbf{Effect of candidate set size $K$}. Increasing $K$ consistently improves navigation performance (Fig.~\ref{fig:kp}). When using a single candidate ($K = 1$), DV-VLN achieves 58 OSR / 43 SR / 39 SPL. Expanding the candidate pool to $K = 2$ yields a clear gain 61 / 47 / 43, and the improvement becomes more substantial at $K = 4$, reaching 65 / 52 / 47, indicating that multi-candidate generation is crucial for recovering from myopic or brittle single-shot decisions. Beyond $K = 4$, the curves exhibit a diminishing-return behavior: $K = 6$ gives 66 / 54 / 48, and $K = 8$ further saturates at 67 / 54 / 49. This suggests that once the candidate pool is sufficiently diverse to include plausible next steps, additional candidates are increasingly redundant and mainly contribute marginal refinement.

\noindent\textbf{Effect of verification trials $P$.} 
We observe a similar saturation trend when increasing $P$ (Fig.~\ref{fig:kp}). With $P = 1$, DV-VLN obtains 60 OSR / 46 SR / 41 SPL; increasing to $P = 2$ improves results to 64 / 49 / 45, and $P = 4$ achieves 65 / 52 / 47, showing that repeated verification stabilizes decision making and reduces variance from stochastic decoding. Further increasing verification to $P = 6$ yields only a small additional gain (67 / 52 / 48), with SR essentially plateauing. This indicates that a moderate number of verification trials already provides a reliable posterior signal for re-ranking, while more trials mainly refine OSR/SPL slightly.

\noindent\textbf{Recommended operating point.} Overall, these results demonstrate a clear trade-off: larger $K$ and $P$ tend to improve performance, yet both curves saturate quickly after a moderate budget. In practice, $K = 4$ and $P = 4$ offers a strong and stable operating point, delivering \textbf{65 OSR / 52 SR / 47 SPL}, while $K \geq 6$ or $P \geq 6$ provides only marginal improvements. This analysis supports that DV-VLN can flexibly adapt to different inference budgets and achieves most of its gains under a moderate verification budget, making the approach suitable for deployment scenarios where latency matters.

\subsection{Case Study}
To qualitatively illustrate how dual verification improves decision reliability, we provide a representative success case in Fig.~\ref{case}. The agent is instructed to enter the bedroom and veer left, then navigate toward the stairs, go downstairs, continue straight, and finally stop in front of the glass pane doors. At the current timestep, the navigation history indicates that the agent has just turned left into an attic-like area with wooden flooring and lights, and the agent observes multiple candidate views with textual descriptions (options A–G).

\noindent\textbf{Single-shot decision can be misled by locally plausible cues.}
As shown in Fig.~\ref{case}, both the original CoT baseline and the single-shot prediction follow a locally consistent but ultimately incorrect hypothesis. The single-shot model predicts “go down the stairs” and selects option D as the supporting view, resulting in an incorrect action. This failure exemplifies a common brittleness of one-pass decision making in VLN: when several candidates share similar semantics (e.g., multiple bedroom-like views), the model may prematurely commit to a plausible-looking option without sufficient consistency checking against the instruction, history, and the full candidate set.

\noindent\textbf{DV-VLN samples alternatives and re-ranks them with interpretable verification signals.} DV-VLN generates multiple CoT-conditioned candidates via sampling decoding and then applies dual verification to each candidate. In Fig.~\ref{case}, DV-VLN produces two representative candidates: one aligns with the single-shot decision (Action: D), while the other proposes Action: B. For each candidate, DV-VLN performs True–False Verification (TFV) and Masked-Entity Verification (MEV) multiple times (here P=4), and computes a verification score by summing the number of successful verifications across both types. The incorrect candidate (Action: D) receives low confidence from verification, with TFV = {FFTF} and MEV = {FFFF}, yielding a score of 1, indicating weak global consistency. In contrast, the correct candidate (Prediction: stairs; Action: B) is strongly supported, with TFV = {TTTF} and MEV = {TTTT}, yielding a substantially higher score of 7. The resulting score gap provides an explicit, human-interpretable rationale for re-ranking, and DV-VLN selects the highest-scoring candidate B, matching the correct navigation decision.

Overall, this case study highlights two practical advantages of DV-VLN: (i) robustness, achieved by delaying commitment through candidate sampling and selecting actions via verification-guided re-ranking; and (ii) interpretability, as the TFV/MEV success patterns and aggregated scores transparently expose why one candidate is preferred over another in a cluttered decision context.

\section{Conclusion}
This paper presents DV-VLN, a verification-guided VLN framework that combines a structured navigational chain-of-thought with dual verification at inference time. DV-VLN samples multiple CoT-conditioned candidate actions and re-ranks them using True–False Verification (TFV) and Masked-Entity Verification (MEV) before execution. Experiments on R2R, RxR (English subset), and REVERIE show consistent gains over strong language-only baselines, with competitive performance against several cross-modal systems. Ablations further verify that TFV and MEV provide complementary improvements and that their combination yields the best results.

Despite these successes, DV-VLN still has limitations. The vision-to-text pipeline may lose fine-grained visual cues, and dual verification adds inference overhead due to repeated sampling and checking. Future work will explore integrating DV-VLN with stronger vision-language models to reduce information loss, and designing more efficient verification schedules (e.g., adaptive budgets) to improve the accuracy–latency trade-off.

\bibliographystyle{IEEEtran}
\bibliography{IEEEabrv,egbib}

\begin{IEEEbiography}[{\includegraphics[width=1in,height=1.25in,clip,keepaspectratio]{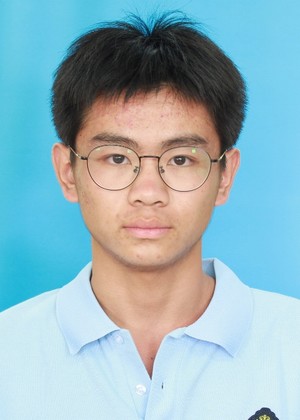}}]{Zijun Li} is currently an undergraduate student majoring in Robotics Engineering at the College of Engineering, Zhejiang Normal University, Jinhua, China. His research interests include vision-and-language navigation, embodied artificial intelligence, and large language model based reasoning.

\end{IEEEbiography}

\begin{IEEEbiography}[{\includegraphics[width=1in,height=1.25in,clip,keepaspectratio]{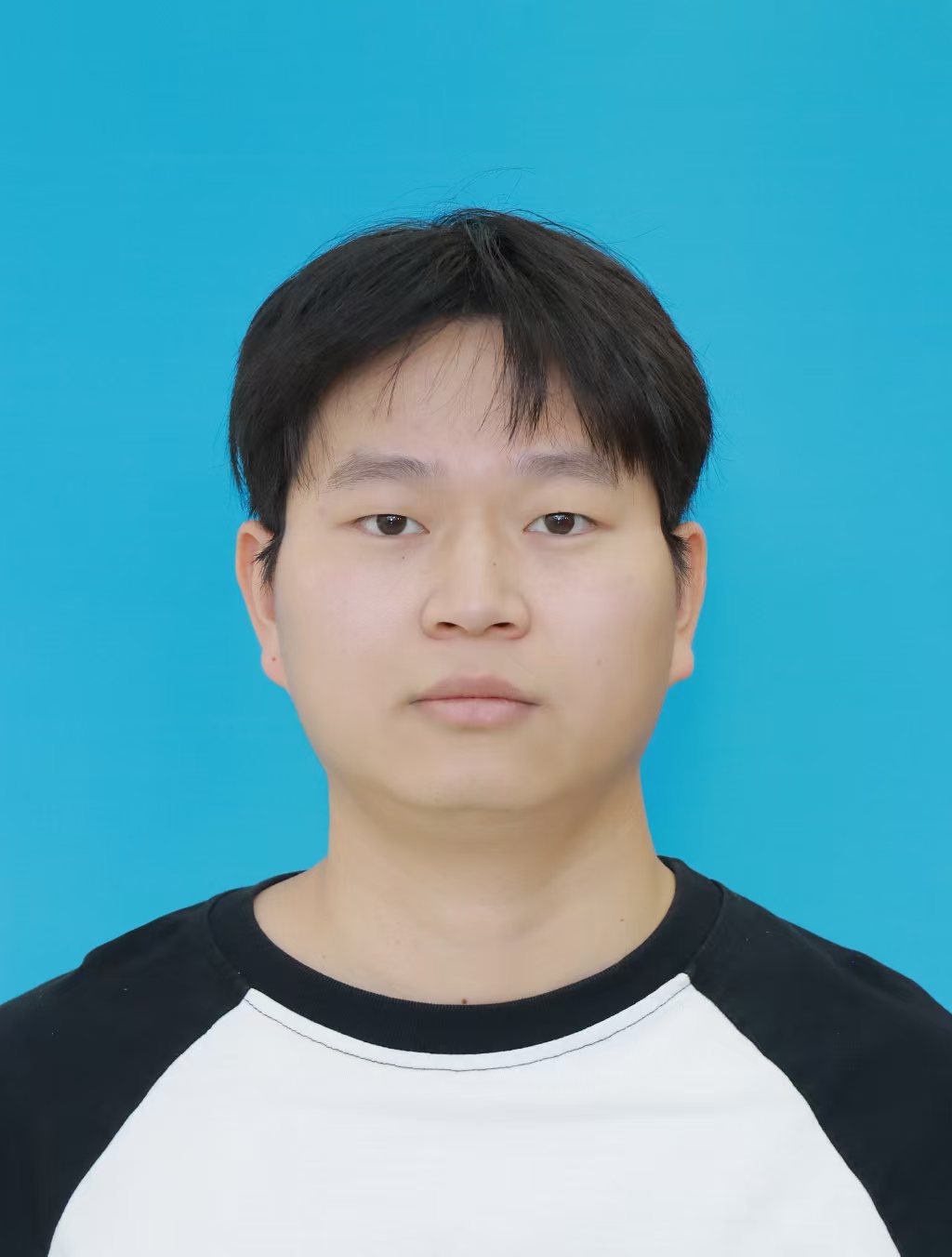}}]{Shijie Li}  
is an Engineer at the Shenzhen Institutes of Advanced Technology (SIAT), Chinese Academy of Sciences. He holds a Master of Science degree and has long focused on the research and development of surgical robot systems based on the ROS (Robot Operating System), as well as related studies such as reinforcement learning. He is deeply dedicated to technological innovation and engineering applications in these fields. He has published 4 academic papers and holds 2 authorized patents in relevant research areas. 
\end{IEEEbiography}

\begin{IEEEbiography}[{\includegraphics[width=1in,height=1.25in,clip,keepaspectratio]{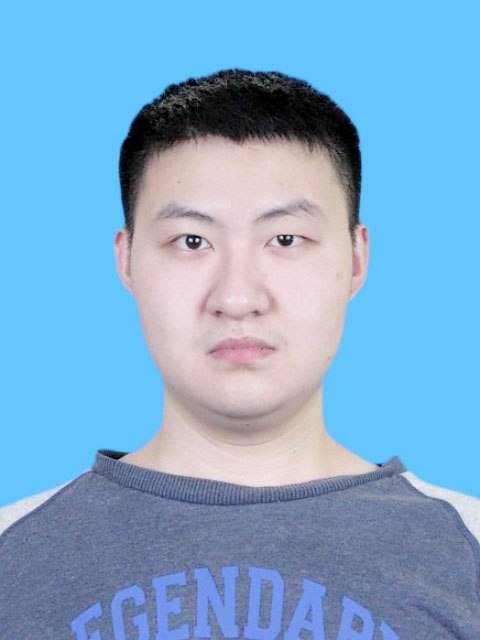}}]{Zhenxi Zhang}

is pursuing his PhD degree in Biomedical Engineering at The Hong Kong Polytechnic University, China. His research focuses on medical image analysis, vision-language models, and embodied robots. Through his work, he aims to advance innovations in healthcare technologies and intelligent systems. 

\end{IEEEbiography}

\begin{IEEEbiography}[{\includegraphics[width=1in,height=1.25in,clip,keepaspectratio]{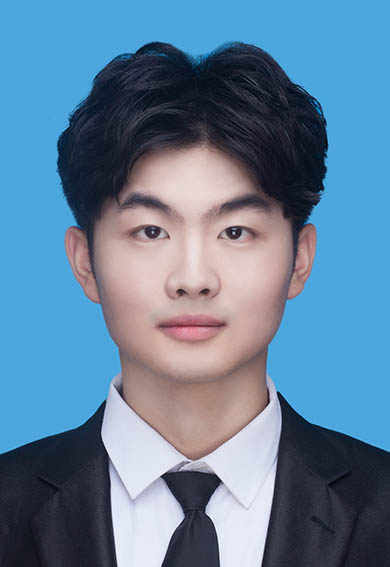}}]{Bin Li}

received the PhD degree from Hunan University, Changsha, China, in 2024. He is now working with the Shenzhen Institutes of Advanced Technology (SIAT), Chinese Academy of Science. His research interests include Vision-and-Language modeling, human-robot natural interaction, and deep learning. 

\end{IEEEbiography}

\begin{IEEEbiography}[{\includegraphics[width=1in,height=1.25in,clip,keepaspectratio]{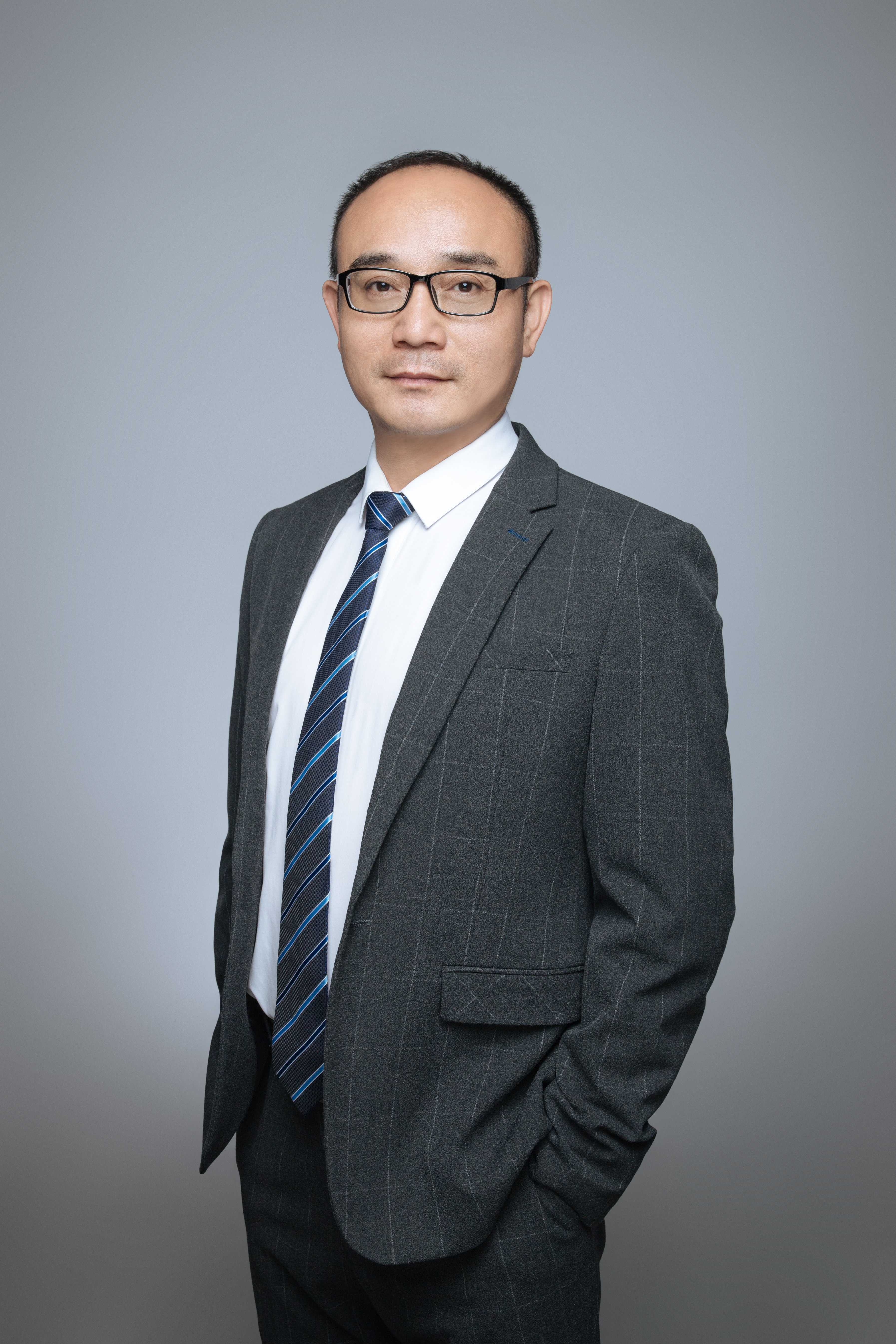}}]{Shoujun Zhou}

Ph.D., is a Research Professor and Doctoral Supervisor at the Shenzhen Institutes of Advanced Technology (SIAT), Chinese Academy of Sciences. His research focuses on the development of intelligent surgical robot systems, with long-term dedication to innovation and application in related fields. He has published 91 academic papers as the first author or corresponding author and holds 22 authorized international patents in relevant research areas.

\end{IEEEbiography}

\end{document}